# BI-RADS prediction of mammographic masses using uncertainty information extracted from a Bayesian Deep Learning model

Mohaddeseh Chegini[a] , Ali Mahloojifar[a]

[a] Department of Biomedical Engineering, Faculty of Electrical and Computer Engineering, Tarbiat Modares University, Tehran, Iran

Correspondence: Ali Mahloojifar (mahlooji@modares.ac.ir)

Abstract

The BI-RADS score is a probabilistic reporting tool used by radiologists to express the level of uncertainty in predicting breast cancer based on some morphological features in mammography images. There is a significant variability in describing masses which sometimes leads to BI-RADS misclassification. Using a BI-RADS prediction system is required to support the final radiologist decisions. In this study, the uncertainty information extracted by a Bayesian deep learning model is utilized to predict the BI-RADS score. The investigation results based on the pathology information demonstrate that the f1-scores of the predictions of the radiologist are 42.86%, 48.33% and 48.28%, meanwhile, the f1-scores of the model performance are 73.33%, 59.60% and 59.26% in the BI-RADS 2, 3 and 5 dataset samples, respectively. Also, the model can distinguish malignant from benign samples in the BI-RADS 0 category of the used dataset with an accuracy of 75.86% and correctly identify all malignant samples as BI-RADS 5. The Grad-CAM visualization shows the model pays attention to the morphological features of the lesions. Therefore, this study shows the uncertainty-aware Bayesian Deep Learning model can report his uncertainty about the malignancy of a lesion based on morphological features, like a radiologist.

## 1. Introduction

Medical decision making is a highly complex probabilistic task and uncertainty plays a fundamental part in Medicine [1, 2]. A radiologist typically reports the findings with some adjectives that reflect the level of uncertainty [3]. One of the probabilistic reporting tools, is the Breast Imaging Reporting and Data System (BI-RADS).

The BI-RADS lexicon was standardized mammographic reporting to facilitate the assessment of mammograms as well as communication between radiologists and referring physicians. The BI-RADS system aims to determine the probability of malignancy and the final recommendation based on specific morphological features described by the shape, margins and density of breast lesions [4]. The shape class is described as oval, round, irregular and lobulated, and the margins are usually circumscribed, obscured, microlobulated, spiculated or ill-defined. Also, the breast density can be classified into four different categories from 1 to 4[5, 6]. Fig.1 illustrates different mass shapes, margins and density described with the BI-RADS lexicon.

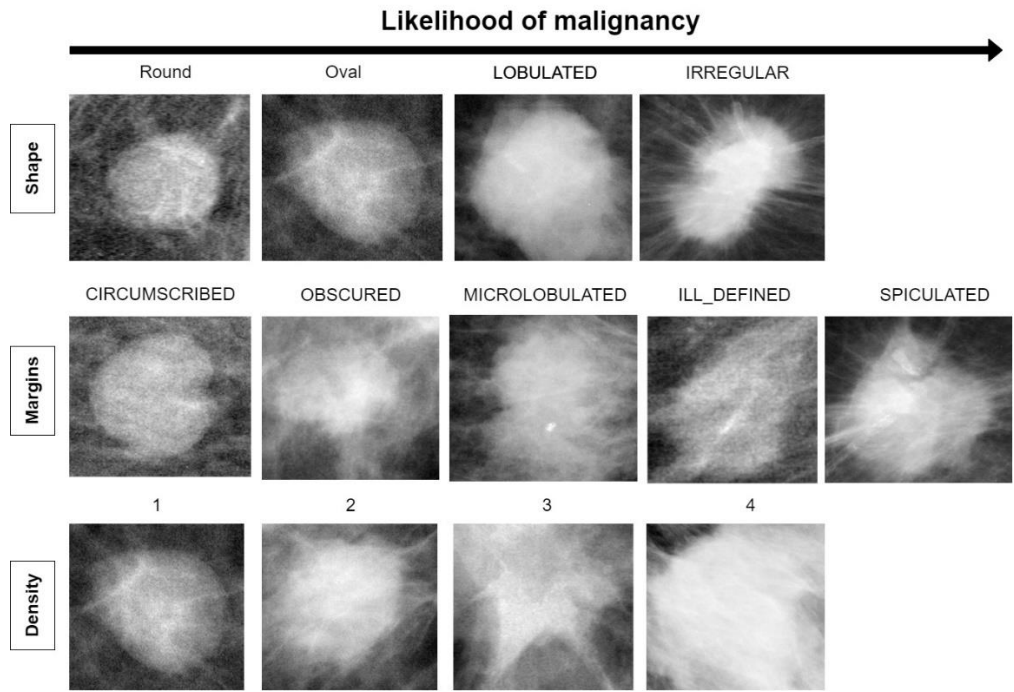

Fig1. Mass lesion description according to the morphological features. The images are some examples from the CBIS-DDSM dataset.

Then, depending on the morphological features of the mass lesion, the radiologists assign one BI-RADS category that is scaled between 0 and 6 indicating the relative probability of malignancy [7]:

(0) BI-RADS 0: incomplete, further imaging evaluations are required.

(1) BI-RADS 1: negative, no abnormality found.

(2) BI-RADS 2: benign.

(3) BI-RADS 3: probably benign.

(4) BI-RADS 4: suspicious finding, likely to be a malignant mass.

(5) BI-RADS 5: highly suggestive of malignancy.

(6) BI-RADS 6: known biopsy-proven malignancy.

For example, a mass with a round or oval shape, circumscribed margin and low density has a high probability of benignity, while a mass with an irregular shape, spiculated margin and high density is suspicious for cancer [5].

In terms of expressing possibilities, BI-RADS 1 indicates a healthy patient, meaning the absence of artifacts in the breast that reveal the presence of cancer, BI-RADS 2 and 5 indicate a high probability of benign and malignant lesions, respectively. BI-RADS 4 subgroup also assigns a wide variability in the probability of malignancy, ranging from 2% to 95%. In fact, BI-RADS 3 and 4 are recommended when there is a high degree of uncertainty for the specialist in selecting the other BI-RADS classes. The details of each BI-RADS category and the probability of malignancy of lesions are shown in Table 1.

Table1. BI-RADS classification

| BI-RADS score | Likelihood of malignancy (%) |
| --- | --- |
| 0 | N/A |
| 1 | 0 |
| 2 | 0 |
| 3 | $0 < p \leq 2$ |
| 4a | $2 < p \leq 10$ |
| 4b | $10 < p \leq 50$ |
| 4c | $50 < p < 95$ |
| 5 | $p \geq 95$ |
| 6 | N/A |

Interpretation of mammograms and BI-RADS description take radiologists an enormous amount of effort and time and it is a challenging task even for experts. On the other hand, there is a large inter-observer variability in describing masses which sometimes leads to BI-RADS classification errors[5, 7, 8]. Thus, using an automatic BI-RADS prediction system is required to support the final decisions and diagnosis of radiologists, as well as help them to reduce the errors, effort and time.

Artificial intelligence (AI) has revolutionized in medical image analysis and has attained many significant achievements[9]. AI aims to build a computer with intelligence that can independently interpret events or think like a person[10].The impressive success achieved by artificial intelligence in healthcare proves that AI can achieve human-like performance [11].

Recently, deep learning (DL) [12], one of the subfields of AI, has played a widespread role in automated medical image analysis[13]. Deep learning models are widely employed to learn features and extract higher-level representations from raw data automatically. A deep learning model is a deep neural network composed of many layers that can transform input data (such as medical images) into outputs (such as predicting the presence of a disease). With the development of deep learning, Convolutional Neural Networks (CNNs), as a variant of deep neural networks, can learn the higher-level predictive features based on imaging examples using multiple layers, including convolution filters inspired by the receptive field in the human primary visual cortex [14]. CNNs have achieved significant success in the field of medical image processing, demonstrating state of the art performance compared to other AI-based algorithms and could even outperform expert humans in some tasks [10].

Many deep learning-based Computer-Aided Diagnosis systems (CADs) proposed for improving the detection of abnormalities, provide good performance and can be a great helpful tool for physicians and experts. Most of the proposed CADs have been used for classifying benign and malignant breast lesions[15-19], while from a clinical aspect, benign/malignant classification disagrees with radiologists' workflow. In fact, radiologists should primarily assign one of the BI-RADS categories to each breast lesion and they cannot claim the malignancy or benignity of the detected abnormality without carrying out a follow-up study or biopsy [7]. Therefore, developing a BI-RADS predictor system remarkably improves the radiologist-AI interaction in the field of breast cancer detection. However, despite its importance, few studies have focused on the design of automatic systems to assign the BI-RADS scores. These studies mainly used machine learning or deep learning models for the conventional classification of BI-RADS, as a multi-class classification task, [20-22], or extract morphological features to describe BI-RADS[5, 7].

On the other hand, despite the advantages, classical deep learning models cannot capture uncertainty [23-25]. In other words, due to the use of point estimates of parameters and predictions in classical deep learning models, these models are too confident about their output and decision without taking into consideration the uncertainty [26, 27]. Many situations can lead to uncertainty in deep learning models. Data uncertainty is arised from ambiguity in the data. This type of uncertainty is called Aleatoric uncertainty or irreducible uncertainty. Also, there is uncertainty about the model, such as selecting the best subset of model parameters or the structure. The model uncertainty is grouped as Epistemic or reducible uncertainty. Epistemic uncertainty can be reduced with more data while aleatoric uncertainty can not. Aleatoric uncertainty and epistemic uncertainty can then be utilized to induce predictive uncertainty[23, 26].

Several different approaches have been proposed to quantify the predictive uncertainty in deep learning models [24]. Compared to all methods, Bayesian-based approaches provide a clear description of the model uncertainty, as well as a deeper theoretical basis[24, 26]. The Bayesian probability theory is used to infer the unobserved quantities, given the observed data. This process be carried out by putting the prior distribution over the space of parameters and combining this information with the likelihood of data to provide posterior distributions [26, 28]. Bayesian deep learning (BDL) uses a combination of the Bayesian probability theory with classical deep learning to provide a predictive probability distribution and extract the uncertainty values.

The Bayesian deep learning models have been widely used in various computer vision tasks, in order to quantify uncertainty. However, as far as we are aware, there has been no previous study that compares the uncertainty of humans (such as a radiologist) and the quantified uncertainty of deep learning models. Because of the lack of ground truth for the uncertainty, the comparison between the estimated and real uncertainty remains as a challenge [24]. On the other hand, it has been shown that the brain may be implementing Bayesian inference [29, 30]. Bayesian probabilistic modeling has emerged not only as the theoretical foundation for rationality in deep learning models, as an artificial intelligence system, but also as a model for normative behavior in humans and animals[26, 31-36]. Indeed, AI, machine learning, the cognitive and the brain sciences have deep common roots. At the cognitive level, these fields converge through Bayesian models of inference [37, 38]. However, as far as we are aware, there has been no previous study that compares and investigates the probabilistic behavior and reported uncertainty of humans and a bayesian deep learning model.

While due to lack of ground truth of the uncertainty, the gap between the estimated and real uncertainty remains as a challenge, there is a reporting tool, namely BI-RADS, in which radiologists use probabilities to reflect their level of uncertainty. Therefore, in this study the uncertainty information extracted by an uncertainty-aware Bayesian deep learning model is used to assign the BI-RADS score, like a radiologist. Also, the predictive probability distribution as well as, the corresponding predictive uncertainty obtained from the bayesian deep learning model are compared with the radiologist, as well as the pathology information.

The main contributions of this study are as follows:

- The uncertainty information extracted from the bayesian deep learning model is used to assign the BI-RADS score to breast lesions, similar to a radiologist.
- The performance of the radiologist versus the model based on the pathology information was investigated in the field of BI-RADS prediction.
- We show that the model considers important areas of the lesion, namely, the shape, margins and density to assign the BI-RADS score.
- This study demonstrates that the uncertainty-aware bayesian deep learning model can infer like a radiologist and report its uncertainty about the malignancy of a lesion based on morphological features.

## 2. Materials and Methods

### 2.1. Dataset

In this study, the CBIS-DDSM (Curated Breast Imaging Subset of DDSM) dataset is used. This dataset provides the ROI (Region Of Interest) corresponding to the lesion in each mammogram image, as well as includes BI-RADS descriptors for mass shape, mass margin and breast density, overall BI-RADS scores from 0 to 5 and rating of the subtlety of the abnormality from 1 to 5. The CBIS-DDSM dataset consists 891 mass cases, including both mediolateral oblique (MLO) and craniocaudal (CC) views of the mammograms. In this study, CC as well as MLO views were used.

The dataset has a defined train and test set that are obtained using 80% and 20% of the cases, respectively. The testing group contains cases of varying difficulty in order to ensure that the method is tested thoroughly.

In this study, each image was resized to $224 \times 224$ pixels and its intensity values were linearly normalized to the interval $[0, 1]$.

Fig.2(a) and Fig.2(b) display the distribution of BI-RADS assessment and pathology labels for the train and test sets. Also, the distribution of data points of BI-RADS assessment with respect to breast density is shown in Fig.2(c).

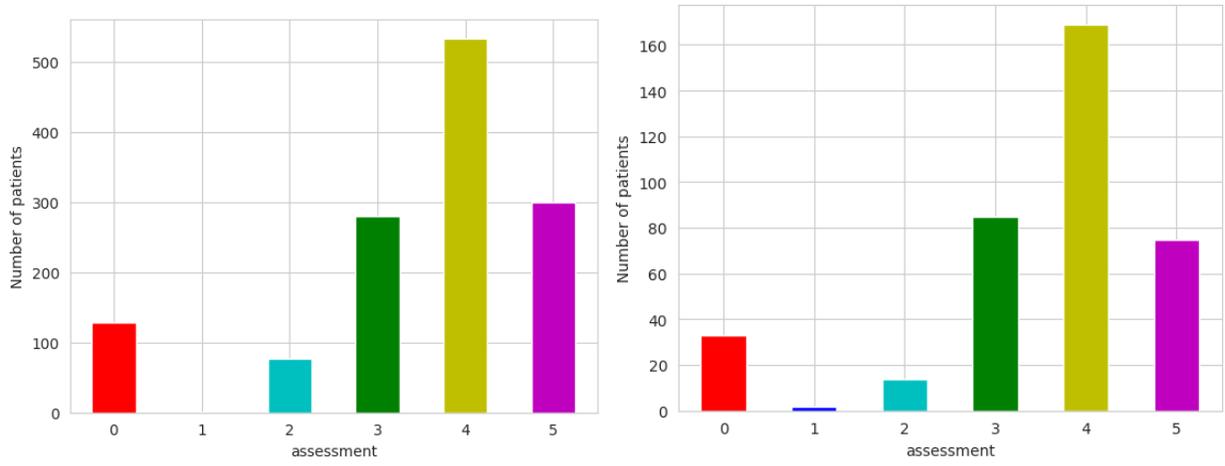

(a)

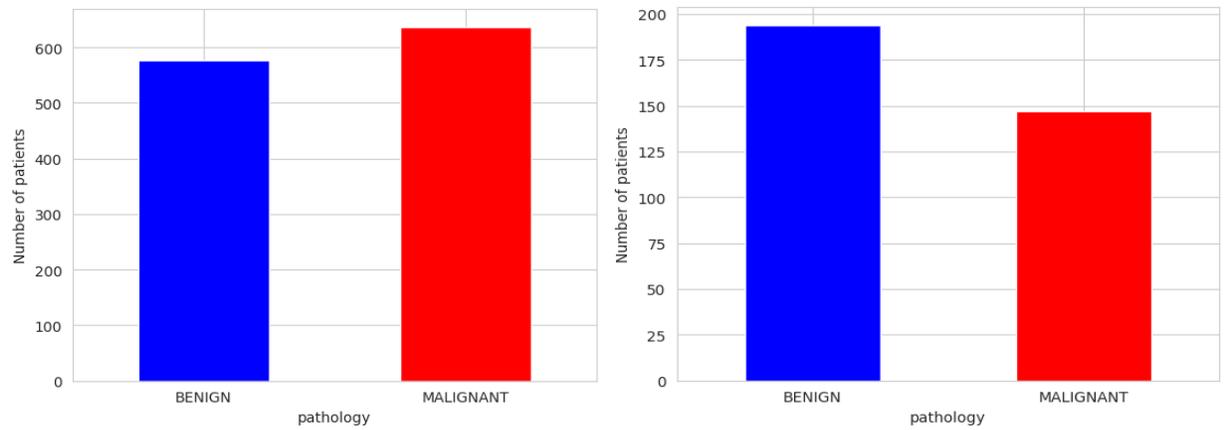

(b)

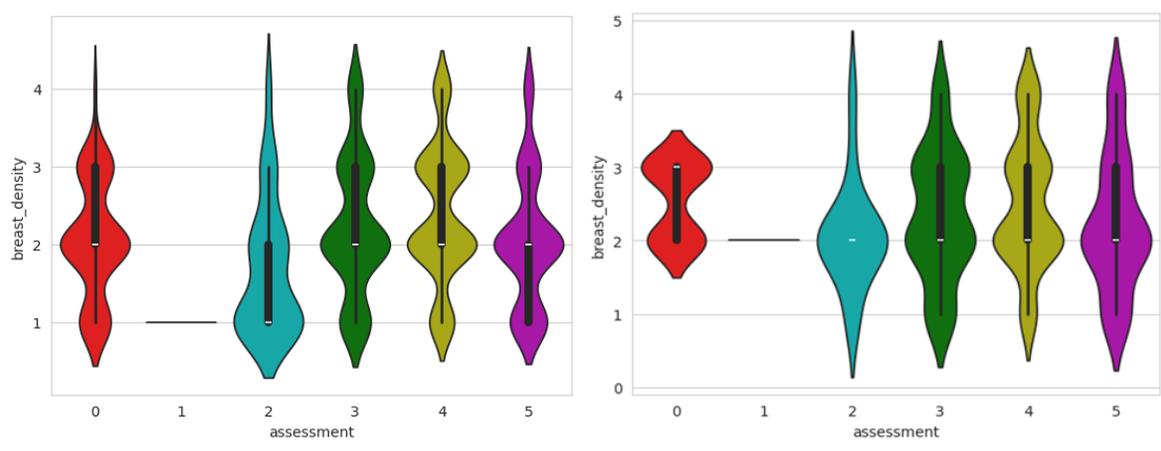

(c)

Fig.2. (a) the distribution of BI-RADS assessment and (b) pathology labels for the train set ( Left column) and the test set (Right column), and (c) distribution of BI-RADS assessment with respect to breast density for the train set ( Left column) and the test set (Right column).

As can be seen in Fig.2(a), there is an imbalance in the different BI-RADS classes in the dataset. In order to overcome the imbalanced number of cases belonging to each BI-RADS category, the SMOTE [39] (Synthetic Minority Over-sampling Technique) method was applied to the dataset.

## 2.2. Bayesian Convolutional Deep Learning model

### 2.2.1. Uncertainty Quantification using Bayesian method

How should an artificially intelligent system represent and update its beliefs about the world in light of data [26]? Almost all Machine Learning, as well as Deep Learning tasks can be formulated as making inferences about the unobserved data from the observed data. The Probabilistic modeling provides a framework for representing beliefs by inferring the models. The probabilistic models can use the beliefs, that learn from data acquired through experience, to make predictions about new data, and make decisions that are rational given these predictions. Since any intelligent model will be uncertain when predicting unobserved data, uncertainty quantification plays a fundamental role in the modeling. A direct application of the law of total probability to compute the basic predictive uncertainty is called Bayes theorem [24, 26].

In order to quantify uncertainty in deep learning models, Bayesian probability theory is combined with classical deep learning to infer when deep models become uncertain. This paradigm is called Bayesian Deep Learning (BDL).

The main idea behind the BDL is to consider a posterior distribution over the space of parameters $w$ given training inputs $X = \{x_1, \ldots, x_N\}$ and their corresponding outputs $Y = \{y_1, \ldots, y_N\}$, instead of a single value. The posterior distribution is then sought using Bayes theorem which is written in eq. (1).

$$p(w|X,Y) = \frac{p(Y|X,w)p(w)}{p(Y|X)}$$

(1)

Using the BDL model, the predictive probability distribution can be calculated for a given test sample $x^*$ by integrating, as shown in eq. (2).

$$p(y^*|x^*,X,Y) = \int p(y^*|x^*,w)p(w|X,Y)dw$$

(2)

This process is called Bayesian inference or marginalization [23, 24]. Although conceptually simple, a fully probabilistic approach to deep learning poses a number of computational challenges. In real-world problems and applications, it is difficult to compute analytically an exact posterior distribution, because of the inability to scale the Bayesian techniques to complex models with potentially millions of parameters and large amounts of data[40].

Therefore, p(w│X,Y) should be approximated by various approximation methods. One of the most widespread approximation methods is variational inference method [41]. According to the variational inference method, in order to find the predictive distribution, an approximating variational distribution q(w) can be defined instead of the true posterior distribution as shown in eq. (3).

$$p(y^*|x^*, X, Y) = \int p(y^*|x^*, w) p(w|X, Y) dw \tag{3}$$

$$\approx \int p(y^*|x^*, w) q(w) dw$$

Many variational inference techniques are computationally expensive when integrated into a deep architecture. To address this problem, the Monte Carlo (MC) dropout approximation method was introduced [42].

Dropout is an effective technique that is used to solve overfitting problems in deep learning models [43]. During the training step, dropout discards some neurons of the Neural Network with a certain probability, known as the dropout rate, for each iteration. This process is called a stochastic forward pass. In the Monte Carlo Dropout method, dropouts can be activated not only during training but also at test time. Therefore, this method utilized T stochastic forward passes during test time and collects them to estimate the predictive distribution using Monte Carlo integration. This process is represented as:

$$p(y^* = c|x^*, X, Y) \approx \frac{1}{T} \sum_t p(y^* = c|x^*, w_t)$$

(4)

Where $w_t \sim q(w)$, $q(w)$ is a Bernoulli variational distribution or a dropout variational distribution and $p(y^* = c|x^*, w)$ is a SoftMax likelihood for a classification task which is written in eq. (5).

$$p(y^* = c|x^*, w) = \frac{\exp(f_c^w)}{\sum_{c'} \exp(f_{c'}^w)} \quad (5)$$

Where $f_c^w$ denotes a network function parametrized by the variables $w$. This means that a network already trained with dropout is a BDL model [40, 42]. The MC dropout is an effective approximation method of Bayesian inference to mitigate the problem of representing uncertainty in deep learning without sacrificing either computational complexity or test accuracy.

Finally, with the predictive distribution $p(y^* = c|x^*, X, Y)$, the predictive uncertainty in test points $x^*$ can be quantified using predictive entropy [40, 44]. Predictive entropy is the entropy of the predictive distribution $p(y^* = c|x^*, X, Y)$, in test points $x^*$, which captures the average information in the predictive distribution. The predictive entropy defined as eq. (6):

$$H_{p(y^*|x^*, X, Y)}[y^*] = -\sum_{y^*=c} p(y^* = c|x^*, X, Y) \log p(y^* = c|x^*, X, Y)$$

(6)

Where $p(y^* = c|x^*, X, Y)$ indicates the Mc dropout approximated predictive distribution obtained by T stochastic forward passes through the model, defined as eq. (4). It should be noted that due to the predictive entropy capturing epistemic and aleatoric uncertainty, this quantity is high when either the aleatoric or epistemic uncertainty is high.

### 2.2.2. Model Architecture

In this paper, the transfer learning technique is applied. Transfer learning is a method that involves reusing a pre-trained CNN model to learn features as the starting point for a model on a new task. Some popular pre-trained CNN models ,that were trained with the ImageNet, are selected and transferred the pre-trained model to the studied datasets for fine-tuning to represent two classes as benign and malignant. In this study, some well-known deep convolutional neural network architectures are used, including: VGG16, VGG19, Xception, ResNet50, ResNet101, ResNet152, InceptionV3, InceptionResNetV2, DenseNet201, EfficientNetB4-5, EfficientNetV2S and EfficientNetV2M.

The classifier blocks include a stack of three Fully-Connected and two MC Dropout layers: the first two Fully-Connected layers have 4096 neurons each with a ReLu activation function, and the third one performs binary classification and thus, includes two neurons with a SoftMax activation function. The MC dropout layer is mainly included for making the Bayesian network to quantify the uncertainty. Therefore, the output of the models includes a prediction about whether the lesion is malignant or benign, in addition to the predictive uncertainty.

### 2.2.3. Learning method

The ADAM optimizer with an initial learning rate of 0.00001 and a batch size of 32 is used for training all models. Also, the number of epochs are 100. The Binary cross-entropy loss function is used to select the optimal weights.

### 2.2.4. Extraction the BI-RADS uncertainty levels

In a classification task, the softmax output indicates the probability score of a class and cannot be associated with model uncertainty. Additionally, the softmax output is often poorly calibrated, leading to inaccurate uncertainty estimates [24]. Therefore, we utilize the predictive entropy derived from the predictive distribution to quantify the uncertainty in a bayesian deep learning model.

On the other hand, the BI-RADS category interpretation is a subjective process for radiologists[3]. The BI-RADS levels represent degree of the predictive uncertainty and determine the predictive probability of malignancy(Table.1).

Therefore, to compare the BI-RADS prediction in the radiologist and the model, the predictive probability of the radiologist and the model, or the corresponding predictive uncertainty can be used. From the perspective of comparing the predictive uncertainties, the BI-RADS scores should be converted into the entropy measures. Therefore, based on the predictive probability of malignancy in an observed lesion, according to Table 1, the predictive entropy values of each BI-RADS level are extracted. Then, we assign a BI-RADS score to each lesion based on the uncertainty information extracted from the model according to the BI-RADS entropy measures. For better understanding, Fig.3 and Fig.4 are presented.

Suppose a lesion is observed by a radiologist and assigned a predictive probability of malignancy, as well as, a probability of benignity. Then, the predictive entropy formula is expressed as follows:

$$H_{p(y^*|x^*,X,Y)}[y^*]$$
$$= - \sum_{y^*=\{malignant,benign\}} p(y^* = c|x^*,X,Y) \log p(y^* = c|x^*,X,Y)$$
(7)

For each BI-RADS category, the entropy measures can be derived according to Table.1, as follows:

(0) BI-RADS 0: -

(1) BI-RADS 1: -

(2) BI-RADS 2:

    Predictive probability:
$$p(y^* = malignant|x^*,X,Y) = 0$$
and
$$p(y^* = benign|x^*,X,Y) = 1 - (y^* = malignant|x^*,X,Y) = 1$$
Corresponding predictive entropy:
$$H_{p(y^*|x^*,X,Y)}[y^*] \cong 0$$
(8)

(3) BI-RADS 3:

    Predictive probability:
$$0 < p(y^* = malignant|x^*,X,Y) < 0.02$$
and
$$0.98 < p(y^* = benign|x^*,X,Y) = 1 - (y^* = malignant|x^*,X,Y) < 1$$
Corresponding predictive entropy:
$$0 \leq H_{p(y^*|x^*,X,Y)}[y^*] \leq 0.14$$
(9)

(4a) BI-RADS 4a:

Predictive probability:
$$0.02 < p(y^* = malignant|x^*, X, Y) < 0.1$$
and
$$0.9 < p(y^* = benign|x^*, X, Y) = 1 - (y^* = malignant|x^*, X, Y) < 0.98$$
Corresponding predictive entropy:
$$0.14 \leq H_{p(y^*|x^*, X, Y)}[y^*] \leq 0.469$$

(10)

(4b) BI-RADS 4b:

Predictive probability:
$$0.1 < p(y^* = malignant|x^*, X, Y) < 0.5$$
and
$$0.5 < p(y^* = benign|x^*, X, Y) = 1 - (y^* = malignant|x^*, X, Y) < 0.9$$
Corresponding predictive entropy:
$$0.469 \leq H_{p(y^*|x^*, X, Y)}[y^*] \leq 1$$

(11)

(4c) BI-RADS 4c:

Predictive probability:
$$0.5 < p(y^* = malignant|x^*, X, Y) < 0.95$$
and
$$0.05 < p(y^* = benign|x^*, X, Y) = 1 - (y^* = malignant|x^*, X, Y) < 0.5$$
Corresponding predictive entropy:
$$0.286 \leq H_{p(y^*|x^*, X, Y)}[y^*] \leq 1$$

(12)

(5) BI-RADS 5:

Predictive probability:

$$0.95 < p(y^* = malignant|x^*, X, Y) < 1$$

and

$$0 < p(y^* = benign|x^*, X, Y) = 1 - (y^* = malignant|x^*, X, Y) < 0.05$$

Corresponding predictive entropy:

$$0 \leq H_{p(y^*|x^*, X, Y)}[y^*] \leq 0.286$$

(13)

(6) BI-RADS 6: -

Now, we can assign a BI-RADS score to each lesion according to the criteria obtained from eq 8 to 13 using extracted uncertainty information by the BDL model.

If the predicted label of the model is benign:

If the predictive entropy is:

$$H_{p(y^*|x^*, X, Y)}[y^*] \cong 0$$

So, the predicted BI-RADS is 2.

If the predictive entropy is:

$$0 \leq H_{p(y^*|x^*, X, Y)}[y^*] \leq 0.14$$

So, the predicted BI-RADS is 3.

If the predictive entropy is:

$$0.14 \leq H_{p(y^*|x^*, X, Y)}[y^*] \leq 0.469$$

So, the predicted BI-RADS is 4a.

And, if the predictive entropy is:

$$0.469 \leq H_{p(y^*|x^*, X, Y)}[y^*] \leq 1$$

So, the predicted BI-RADS is 4b.

In contrast, if the predicted label of the model is malignant:

If the predictive entropy is:

$$0.286 \leq H_{p(y^*|x^*, X, Y)}[y^*] \leq 1$$

So, the predicted BI-RADS is 4c.

And, if the predictive entropy is:
$$0 \leq H_{p(y^*|x^*,X,Y)}[y^*] \leq 0.286$$

So, the predicted BI-RADS is 5.

Where $p(y^*|x^*,X,Y)$ indicates the predictive distribution obtained from the BDL model, $x^*, y^*$ is a test point and corresponding label, X,Y are training inputs, and their corresponding outputs.

For better understanding, the predicted labels, the predictive probability values and the corresponding predictive entropy are shown in Fig.3. Also, summary of this study is shown in Fig.4.

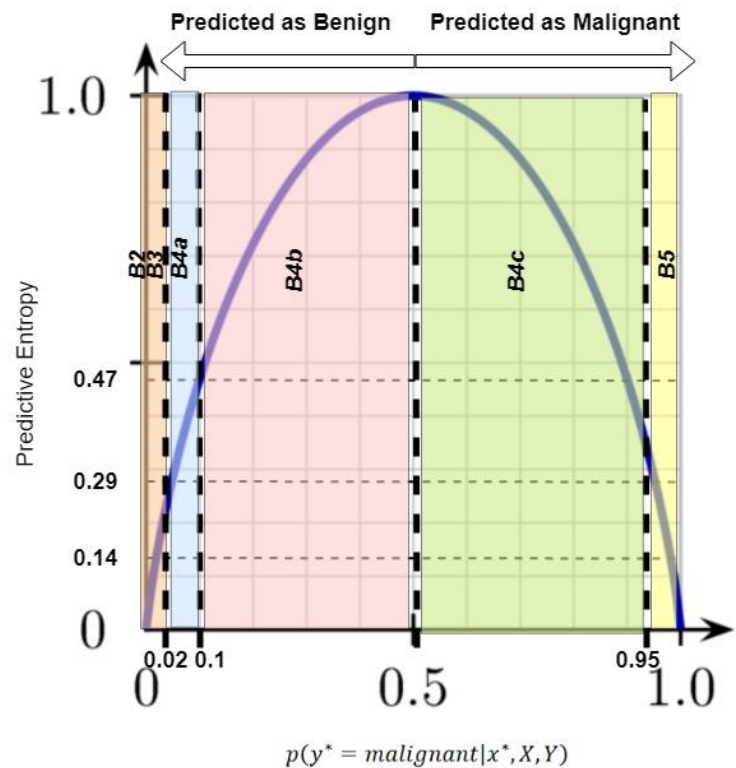

Fig.3. The predictive probability of malignancy and the corresponding predictive entropy. The threshold for assigning each BI-RADS score is specified by the BI-RADS standard lexicon (Table 1). For example, suppose the BDL model predicts a sample as a benign lesion with an uncertainty of 0.3, which, according to the BI-RADS criterion, we assign to this sample a BI-RADS score of 4a. Our goal with this thresholding is alignment the BDL model and the radiologist. (B2: BI-RADS 2, B3: BI-RADS 3, B4a: BI-RADS 4a, B4b: BI-RADS 4b, B4c: BI-RADS 4c, B5: BI-RADS 5)

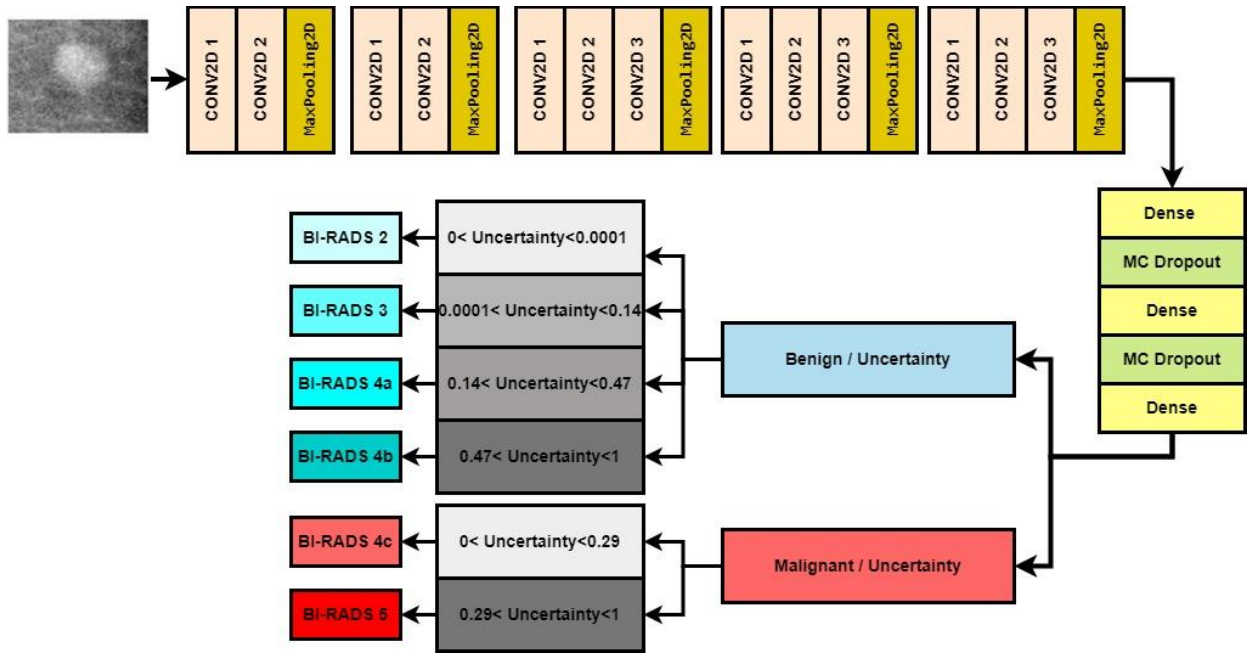

Fig.4. A general overview of the proposed bayesian deep learning model to predict the BI-RADS score based on the uncertainty information.

## 3. Results

### 3.1. Experimental setup

To implement the models, Python 3, TensorFlow and Keras framework are used. All the experiments have been executed on the NVIDIA Tesla P100 GPU with 16 GB memory configured machine.

### 3.2. Radiologist performance

Typically, in a BI-RADS prediction and classification task, the predicted classes are evaluated with the label that a radiologist assigns to a mammography images. In this section, the purpose is to check the acuracy of the BI-RADS labels of the radiologist.

As mentioned, in the field of decision making in BI-RADS description and prediction using mammography, there exists significant variability among radiologists. In a study, based on deep learning assistance, the participants changed their original BI-RADS classification on average $21.3 \pm 2.5$ times [45]. Therefore, radiologists may make mistakes in assigning BI-RADS to mammography images.

To demonstrate the accuracy of the radiologist's BI-RADS prediction, the BI-RADS labels are compared to the binary benign/malignant label confirmed by pathology information. In fact, all benign cases are expected to belong to one of the B2, B3, or at least B4a and B4b classes. Conversely, all malignant cases must belong to one of the B4C and B5 classes. It should be noted that due to the lack of conventional labeling of the B4 class into subgroups 4a, 4b, 4c in the used dataset, we can only examine the B2, B3, and B5 classes with corresponding pathology labels.

The investigation results show that in BI-RADS 2 class, the radiologist's prediction accuracy is 75%. Similarly, the prediction accuracy in B3 and B5 classes was 93.55% and 93.33%, respectively. It should be noted that the misclassification results occur due to assign B2 and B3 labels to malignant samples and B5 to benign samples. Also, the radiologist's f1-scores are 42.86%, 48.33% and 48.28% in the BI-RADS 2, 3 and 5 classes, respectively

### 3.3. Model performance

As previously stated, the radiologist's label may not be correct. Therefore, for each BI-RADS class, a comprehensive comparison is separately performed between the prediction of the radiologist and the model, considering pathology labels. The goal is comparison between the model and radiologist performance in a fair way.

1. Category BI-RADS 2:

This class in the CBID-DDSM test dataset consists of 4 samples. Three samples are benign and one sample is malignant, confirmed by pathology information. Table 2 and Table 1 in the supplementary file examine in detail the performance of the model and the radiologist in predicting these samples.

As can be seen in Table 1 in the supplementary file, the predicted BI-RADS of model differs from the radiologist in some cases. In one case, the model correctly assigned a BI-RADS 5 label to a malignant case, while the radiologist assigned a BI-RADS 2 label.

As mentioned before, all benign cases are expected to belong to one of the B2, B3, or at least B4a and B4b classes. Conversely, all malignant cases must belong to one of the B4C and B5 classes. Accordingly, the performance of the radiologist and the BDL model are shown in Table 2. The results show while the model and radiologist accuracy are equal, the f1-score of the model is better than the radiologist.

Table2. Performance (%) of the radiologist versus the BDL model based on pathology information in the BI-RADS 2 category.

| | Pathology Label | Precision | Recall | f1-score | Accuracy |
|---|---|---|---|---|---|
| Radiologist | Benign | 75 | 100 | 85.71 | |
| | Malignant | 0 | 0 | 0 | |
| | (macro)Average | 37.50 | 50.00 | 42.86 | 75 |
| Model | Benign | 100 | 66.67 | 80.00 | |
| | Malignant | 50.00 | 100 | 66.67 | |
| | (macro)Average | 75.00 | 83.33 | 73.33 | 75 |

2. Category BI-RADS 3:

This class in the CBIS-DDSM test dataset consists of 62 samples. 58 samples are benign and 4 samples are malignant, confirmed by pathology information. Table 3 and Table 2 in the supplementary file examine in detail the performance of the model and the radiologist in predicting these samples.

The results obtained from Table 3 show that the accuracy of the model based on pathology information is 79.03%. While the accuracy of the radiologist is 93.55%. Also, the f1-score of the model is 59.60%. This measure is 48.33% for the radiologist.

The model classified 26 samples from 58 benign samples in this category as B2. Actually, in practice, there is significant interobserver variability in the assessment of BI-RADS 3 findings. Failure to diagnose correctly may lead to increase healthcare costs, unnecessary radiation exposure, and undue patient anxiety [46]. In this sense, this may be an advantage for the model.

As another advantage, the four malignant samples that the radiologist predicted them as B3, the model correctly predicted three of them as B5 and B4c classes. Examination of the incorrectly predicted samples shows that the shape, margins or density of the lesion may mislead the model. According to Table 2 in the supplementary file, most misclassified samples at least have the shape and margins or density close to the malignancy were predicted as B4 or B5. Features that caused the model make incorrect predictions as B4 or B5 are highlighted yellow in Table 2 in the supplementary file.

Table3. Performance (%) of the radiologist versus the BDL model based on pathology information in the BI-RADS 3 category.

| | Pathology Label | Precision | Recall | f1-score | Accuracy |
|---|---|---|---|---|---|
| | Benign | 93.55 | 100 | 96.67 | |
| Radiologist | Malignant | 0 | 0 | 0 | |
| | (macro)Average | 46.77 | 50.00 | 48.33 | 93.55 |
| Model | Benign | 97.87 | 79.31 | 87.62 | |
| | Malignant | 20.00 | 75.00 | 31.58 | |
| | (macro)Average | 58.94 | 77.16 | 59.60 | 79.03 |

3. Category BI-RADS 4:

The BI-RADS 4 subgroup assigns wide variability in the probability of malignancy, ranging from 2% to 95%[47]. In fact, the BI-RADS 4 category occurs when there are overlapping morphological features of benign and malignant lesions, extracted using standard mammography. In the test samples of the used dataset, 165 samples were predicted as BI-RADS 4 by the radiologist, in which 96 samples are benign and 69 samples are malignant, confirmed by pathology information. Table 4 and Table 3 in the supplementary file examine in detail the performance of the model and the radiologist in predicting these samples.

The results obtained from Table 4 show that the accuracy of the model based on pathology information is 68.48% and the f1-score is 67.33%. It is worth noting that, as previously stated, it is not possible to evaluate the performance of B4 prediction of the radiologist with respect to pathology information, because B4 class was not divided into subgroups 4a, 4b, and 4c.

Table 4. Performance (%) of the BDL model based on pathology information in the BI-RADS 4 category.

| | Pathology Label | Precision | Recall | f1-score | Accuracy |
|---|---|---|---|---|---|
| Radiologist | Benign | - | - | - | - |
| | Malignant | - | - | - | |
| | (macro)Average | - | - | - | - |
| Model | Benign | 72.00 | 75.00 | 73.47 | |
| | Malignant | 63.08 | 59.42 | 61.19 | |
| | (macro)Average | 67.54 | 67.21 | 67.33 | 68.48 |

4. Category BI-RADS 5:

The used dataset contains 75 test samples that the radiologist predicted them as BI_RADS 5. According to the pathology information, 5 samples are benign and 70 samples are malignant. Table 5 and Table 4 in the supplementary provide a detail overview of the radiologist and the model performance in predicting these samples.

The results show that the accuracy of the model based on pathology information is 85.33%. Meanwhile the accuracy of the radiologist is 93.33%. The f1-score of the model is 59.26% whereas this measure is 48.28% for the radiologist.

Table5. Performance (%) of the radiologist versus the BDL model based on pathology information in the BI-RADS 5 category.

|  | Pathology Label | Precision | Recall | f1-score | Accuracy |
|---|---|---|---|---|---|
| Radiologist | Benign | 0 | 0 | 0 |  |
|  | Malignant | 93.33 | 100 | 96.55 |  |
|  | (macro)Average | 46.67 | 50.00 | 48.28 | 93.33 |
| Model | Benign | 20.00 | 40.00 | 26.67 |  |
|  | Malignant | 95.38 | 88.57 | 91.85 |  |
|  | (macro)Average | 57.69 | 64.29 | 59.26 | 85.33 |

5. Category BI-RADS 0

Category BI-RADS 0 means the findings are unclear and the radiologist will need more images to determine a BI-RADS score.

The CBIS-DDSM dataset contains 29 samples that were predicted BI-RADS 0 by the radiologist. In this category, 26 samples are benign and 3 samples are malignant, confirmed by pathology information.

While the radiologist would not be able to give any other Bi-RADS scores in these cases, the model was able to distinguish malignant from benign samples with an accuracy of 75.86% and the f1-score is 65.30% (Table6). Also, the model was able to correctly identify all malignant samples in this category. All of the samples that were incorrectly identified also had characteristics such as the shape, margins, or density of the lesion close to malignancy which are highlighted yellow in Table 5 in the supplementary.

Table6. Performance (%) of the BDL model based on pathology information in the BI-RADS 0 category.

|  | Pathology Label | Precision | Recall | f1-score | Accuracy |
|---|---|---|---|---|---|
| Radiologist | Benign | - | - | - |  |
|  | Malignant | - | - | - |  |
|  | (macro)Average | - | - | - | - |
| Model | Benign | 100 | 73.08 | 84.44 |  |
|  | Malignant | 30.00 | 100 | 46.15 |  |

| | (macro)Average | 65.00 | 86.54 | 65.30 | 75.86 |
|---|---|---|---|---|---|

Finally, the distribution of assigned BI-RADS score by the radiologist and the model in the benign and malignant samples are examined. The Fig.5 shows the model tends to use less BI-RADS 4 than the radiologist.

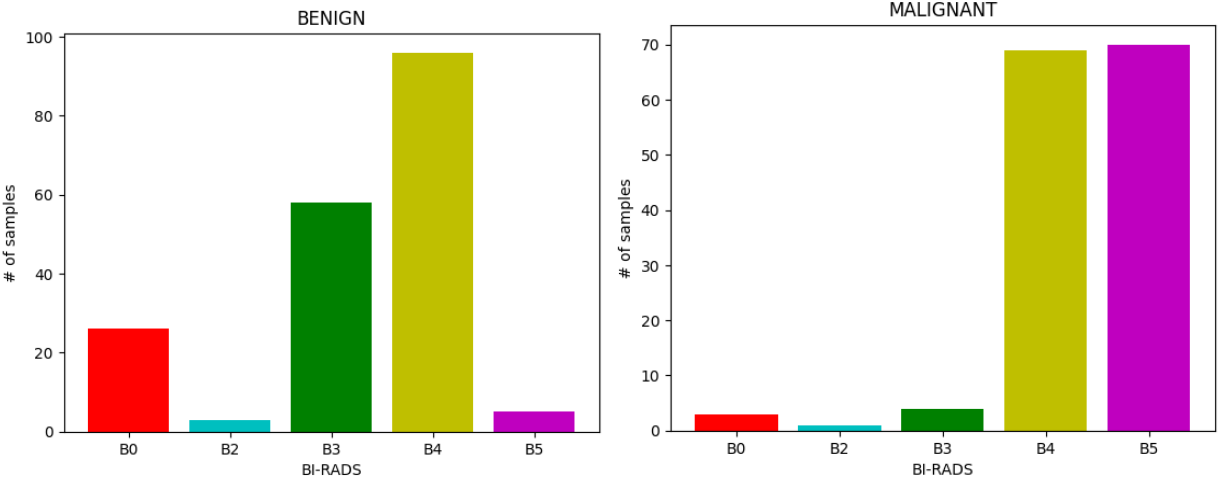

(a)

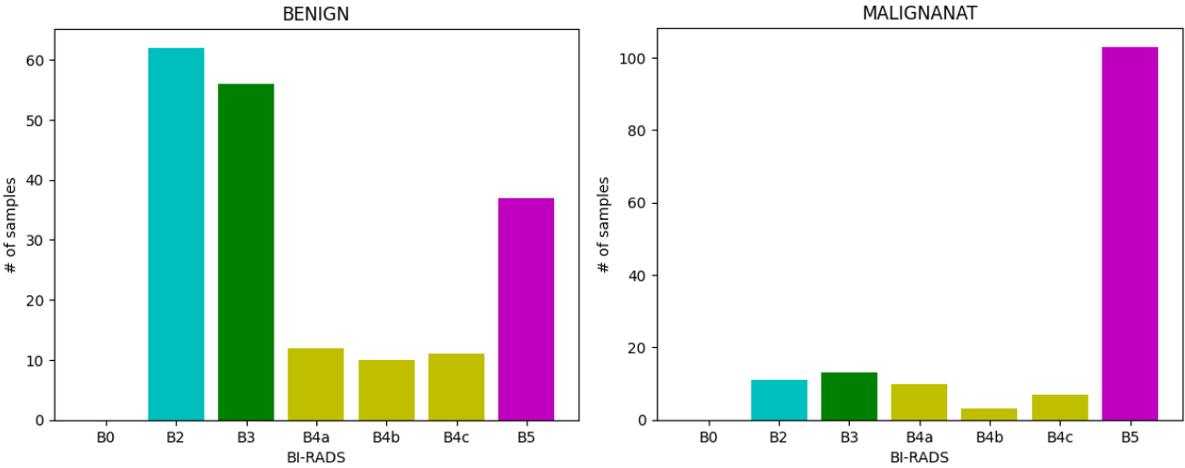

(b)

Fig5. The distribution of assigned BI-RADS score by (a) the radiologist and (b) the BDL model in the benign (Left column) and malignant (Right column) samples.

3.4. Predicted BI-RADS by the radiologist versus the model

Evaluation of breast cancer on mammography to assign a BI-RADS score is a highly complex task due to dependency on numerous factors such as visual perceptions, clinical judgment or the experience of the interpreter. So, there is significant variability among radiologists to assign a BI-RADS score [2].

In this study, we proposed an uncertainty-aware Bayesian deep learning model and attempted to assign a BI-RADS score based on the predicted uncertainty, similar to radiologists.

In the previous section, the predicted BI-RADS score obtained by the model and the radiologist was assessed using the pathology information. In this section, the similarity comparison of the predicted BI-RADS scores by the radiologist versus the model is investigated.

As the confusion matrix in Fig.6 shows, although the BI-RADS predictions of the model and the radiologist differ in many samples, there is a great similarity in assigning BI-RADS 5 to malignant samples.

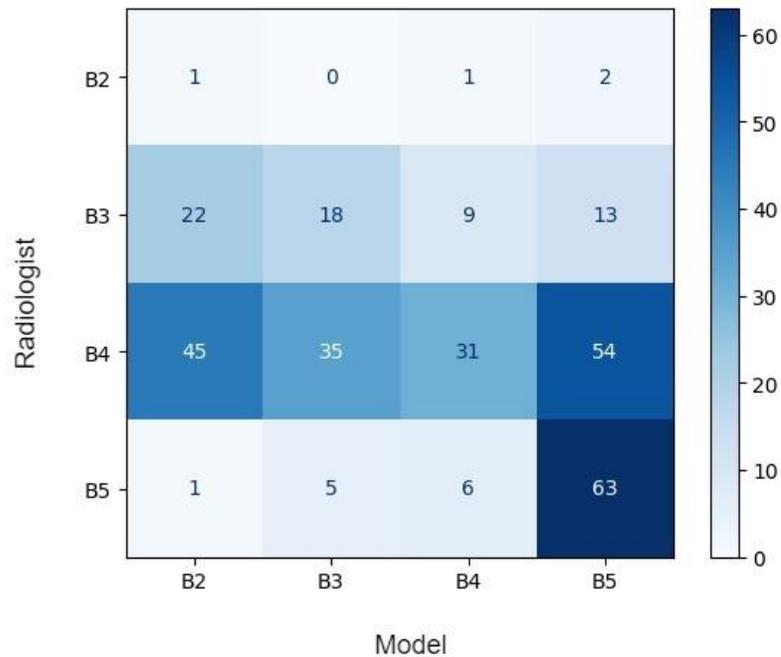

Fig.6. Confusion matrix to compare the similarity between the predicted BI-RADS by the radiologist versus the model

### 3.5. Visual Explanation of important regions

As mentioned, the assignment of BI-RADS to a breast lesion is determined by the morphological features of the lesion, such as the shape, margins, and density. Therefore, it is important that the model also considers important areas of the lesion, namely, the shape, margins and density. So, to investigate the decisions made by the model, the extracted features from the final convolutional layer are used to generate a heatmap via the Grad-CAM visualization procedure. Using the Grad-CAM technique also enhance the interpretability of the models by producing visual explanations for decisions.

Fig.7 presents Grad-CAM visualizations for correctly predicted BI-RADS. These figures show Grad-CAM of some benign samples labeled as B2, B3, B4a, B4b and some malignant cases with B4c, B5 labels. The radiologist's BI-RADS score, predicted BI-RADS of the model and pathology information are written above each sample.

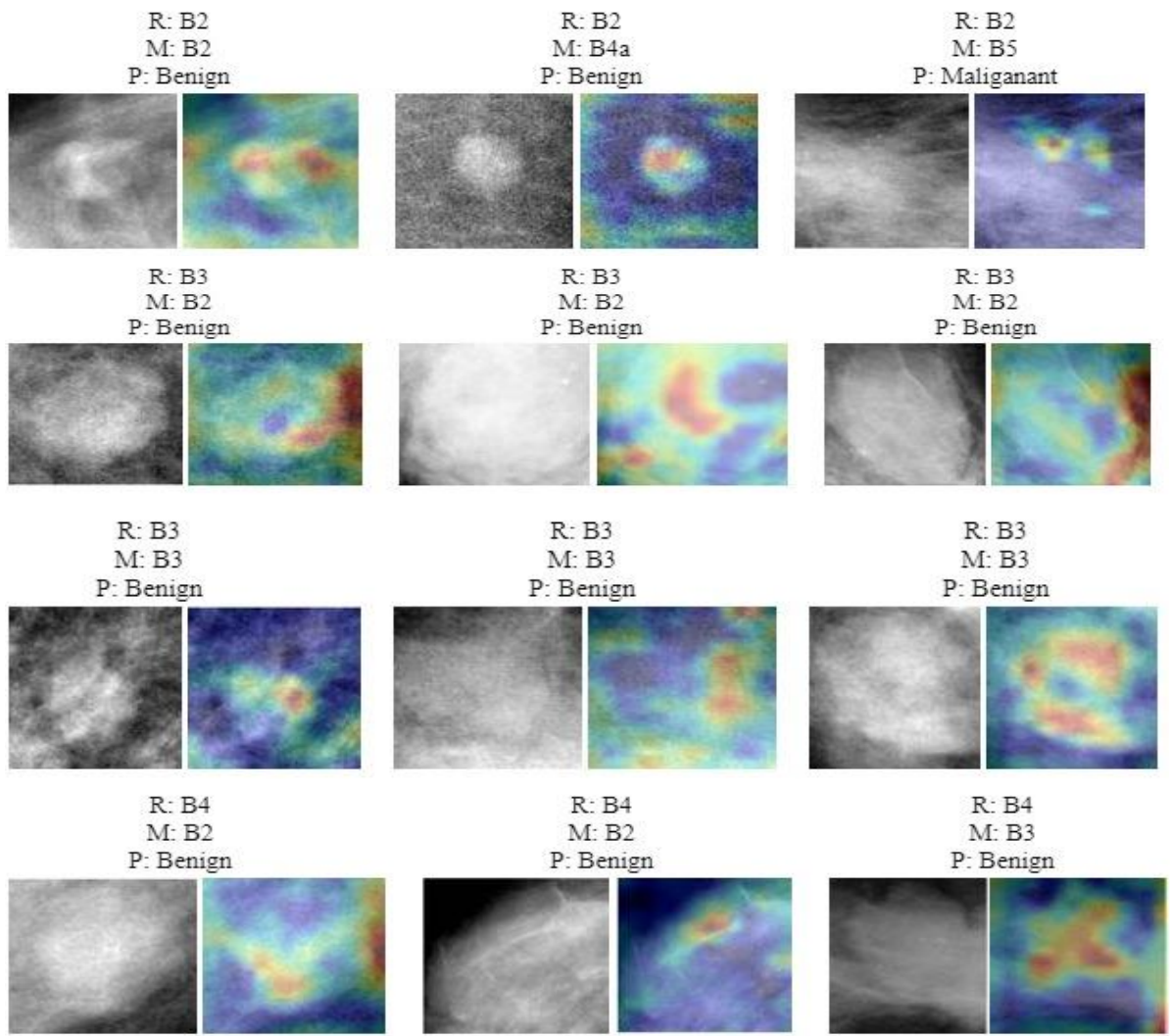

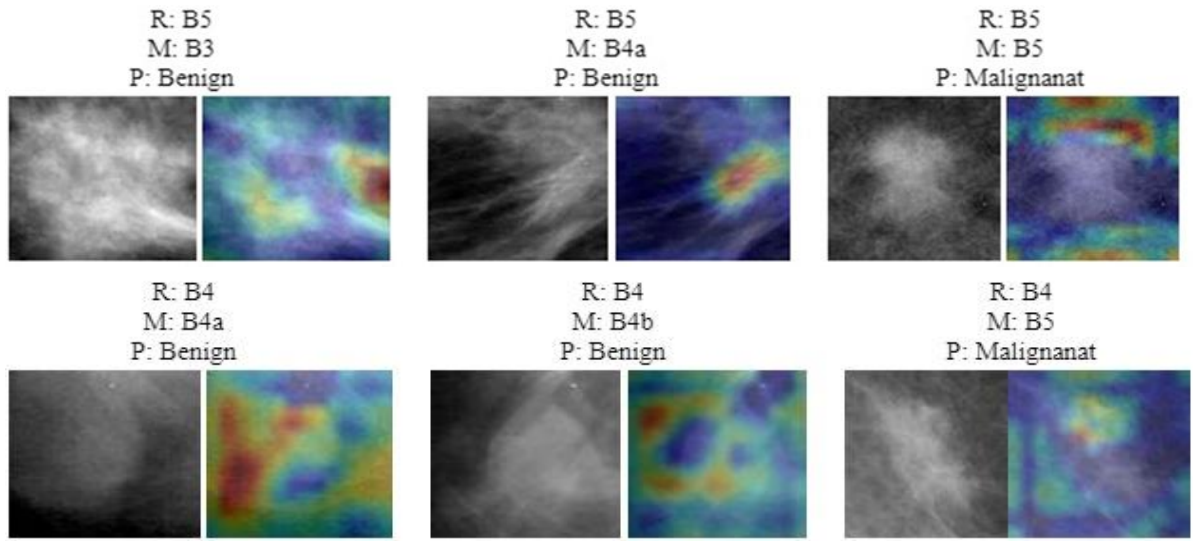

Fig.7. Grad-CAM visualizations for some samples with correctly predicted BI-RADS

Also, Fig. 8 presents Grad-CAM visualizations for misclassified samples. These figures show Grad-CAM of some benign samples labeled as B4c or B5 and some malignant cases with B2, B3, B4a or B4b labels. The radiologist's BI-RADS score, predicted BI-RADS of the model and pathology information are written above each sample.

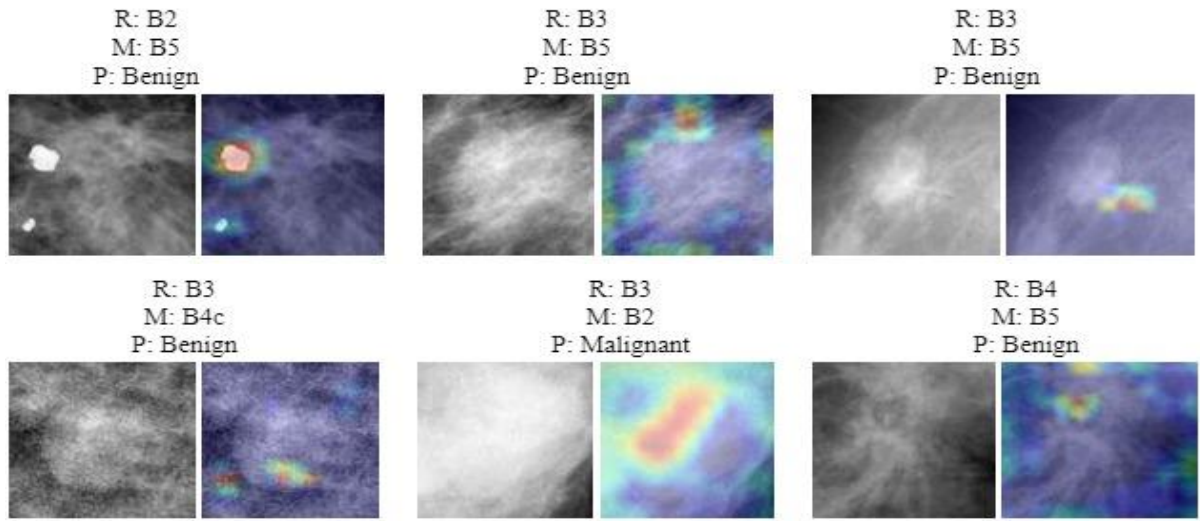

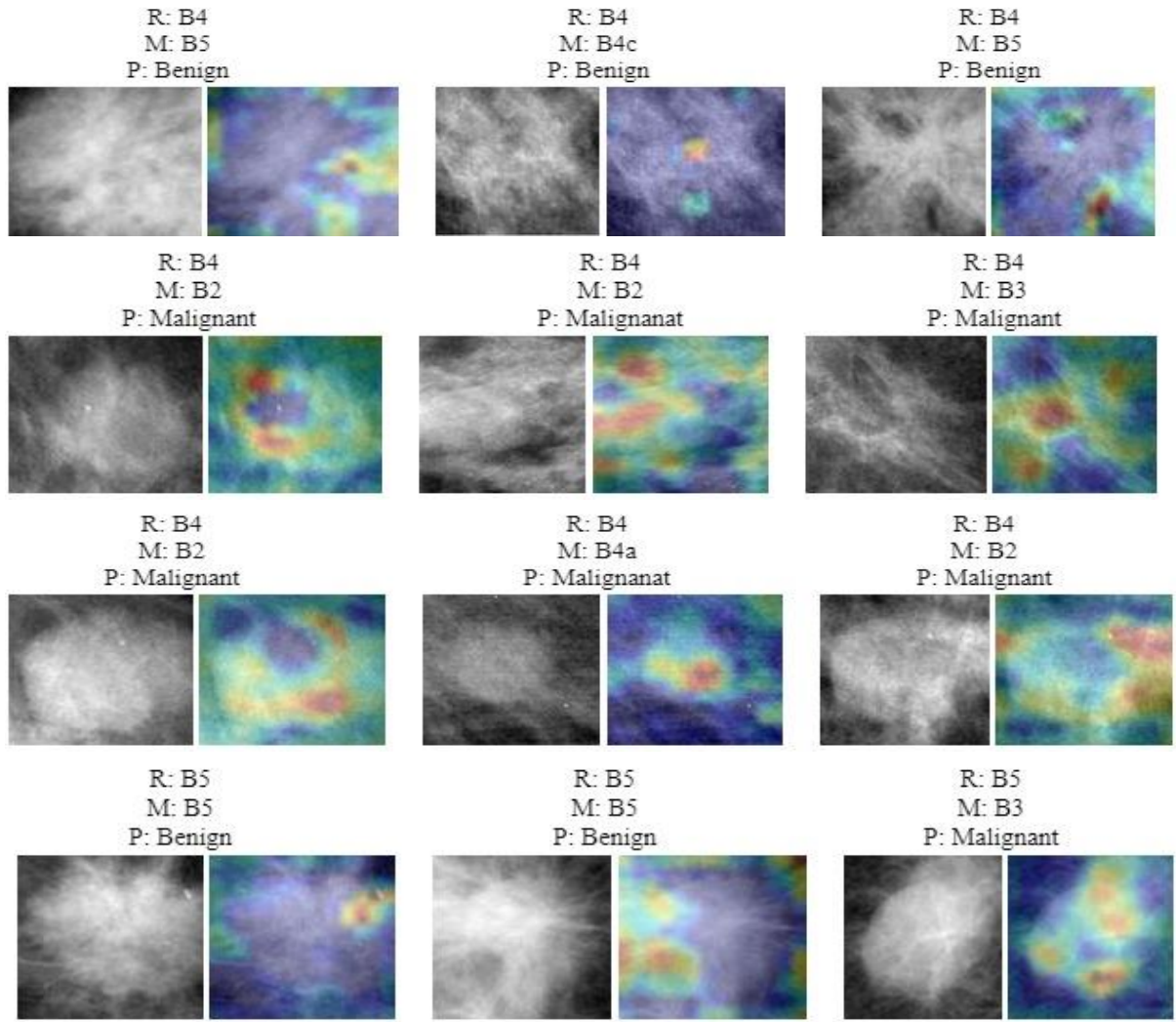

Fig.8. Grad-CAM visualizations for some samples with incorrectly predicted BI-RADS

From Fig 7 and 8, it seems that the model pays attention to the shape, margins and density. Meanwhile, due to the similarity of the morphological characteristics of benign lesions to malignancy, or vice versa, the model occasionally makes mistakes in predicting the type of lesion and the corresponding BI-RADS score.

### 3.6. Ablation study: Model selection

This section presents the ablation study on the some existing pre-trained CNN models to classify breast lesions. The pre-trained model are transferred and fine-tuned the entire model by the used datasets. It is worth noting that the Monte Carlo Dropout layers are used in the classifier blocks of all models to convert the CNN models into Bayesian deep learning models.

First, the performance of some popular existing pre-trained convolutional architectures, namely VGG16, VGG19, Xception, ResNet50, ResNet101, ResNet152, InceptionV3, InceptionResNetV2, DenseNet201, EfficientNetB4-5, EfficientNetV2S and EfficientNetV2M are investigated for breast cancer detection from mammography images, with the same set of hyperparameters. The performances of the pre-trained networks on the dataset in terms of accuracy, recall, precision and f1-score are shown in Table7. Among the pre-trained CNN models, the VGG16 model achieved slightly higher performance in classification of malignant and benign lesions. Therefore, we consider the Bayesian-VGG16 model as the proposed model to predict the BI-RADS score using the uncertainty information.

Table7. Performance (%) of some CNN models.

| Model | Accuracy | Precision | Recall | F1-score |
|---|---|---|---|---|
| VGG16 | 74.48 | 74.23 | 74.48 | 74.29 |
| VGG19 | 72.40 | 72.17 | 72.41 | 72.21 |
| Xception | 73.29 | 72.93 | 72.79 | 72.85 |
| ResNet50 | 69.44 | 69.03 | 69.06 | 69.04 |
| ResNet101 | 66.17 | 66.90 | 66.97 | 66.17 |
| ResNet152 | 73.00 | 72.63 | 72.53 | 72.57 |
| InceptionV3 | 72.70 | 72.65 | 72.96 | 72.59 |
| InceptionResNetV2 | 70.62 | 70.48 | 70.75 | 70.47 |
| DenseNet201 | 72.40 | 72.93 | 70.81 | 71.01 |
| EfficientNetB4 | 70.62 | 70.22 | 69.91 | 70.01 |
| EfficientNetB5 | 72.40 | 72.32 | 72.62 | 72.29 |
| EfficientNetV2S | 71.81 | 71.43 | 71.46 | 71.45 |
| EfficientNetV2M | 70.33 | 71.10 | 71.18 | 70.32 |

4. **Discussion and Conclusion**

Breast cancer detection using mammogram images is a complex probabilistic task due to the dependency on numerous factors such as visual perceptions, clinical judgment or the experience of interpreter. Hence, uncertainty plays a fundamental role in this process. To facilitate reporting the level of uncertainty in breast cancer prediction, radiologists typically use the BI-RADS lexicon. The BI-RADS score is utilized to determine the probability of malignancy of an observed breast lesion by some morphological features, namely, the shape, margins and density.

Interpretation of mammograms and BI-RADS prediction take radiologists an enormous amount of effort and time, meanwhile, there is large variability in describing masses which can occasionally lead to misclassify the BI-RADS scores. Failure to correct diagnose the BI-RADS score in benign lesions may lead to increase healthcare costs, unnecessary radiation exposure, and undue patient anxiety. Conversely, misclassifying of the BI-RADS in malignant lesions can also decrease the treatment chances and increase the mortality rate. Therefore, the use of an automatic BI-RADS prediction system is necessary to support the final radiologist decisions.

Deep learning models have played a widespread role in automated medical image analysis through CAD systems. However, many of deep learning-based CAD systems are proposed for classifying benign and malignant breast lesions, meanwhile benign/malignant classification disagrees with the radiologist workflow in which they assign one of the BI-RADS categories to each breast lesion using mammograms. Consequently, the development of a BI-RADS prediction system remarkably improves the agreement among radiologist-AI interaction in the field of breast cancer detection.

Despite its importance, few studies have focused on the design of an automatic system to assign the BI-RADS scores. These studies mainly used machine or deep learning models for the conventional classification of BI-RADS, as a multi-class classification task, or extract morphological features to describe BI-RADS. While BI-RADS prediction is an uncertainty-dependent task in the diagnosis of malignancy of a lesion, no study has used the uncertainty information obtained from a deep learning model, to predict the BI-RADS score, similar to a radiologist. Therefore, in this study, the uncertainty information extracted by an uncertainty-aware Bayesian deep learning model is used to design an automatic BI-RADS score system. We compare the predictive uncertainty obtained from the bayesian deep learning model with the radiologist, as well as the pathology information. A summary of the investigation results is provided as follows.

- In this study, first, the accuracy of the BI-RADS labels of the model versus the radiologist was investigated. The investigation process is based on the pathology information that determines the benignity/malignancy of a lesion. In fact, all benign cases are expected to belong to one of the B2, B3, or at least B4a and B4b classes. Conversely, all malignant cases must belong to one of the B4C and B5 classes. It should be noted that due to the lack of conventional labeling of the B4 class into subgroups 4a, 4b, 4c in the used dataset, we can only examine the radiologist performance in the prediction of the B2, B3, and B5 classes with corresponding pathology labels. The investigation results show that in the BI-RADS 2, 3 and 5 classes, the radiologist's prediction accuracy are 75%, 93.55% and 93.33%, respectively. On the other hand, the prediction accuracy of the model are 75%, 79.03%, 68.48% and 85.33% in the BI-RADS 2, 3, 4 and 5 classes, respectively. Also, the radiologist's f1-scores are 42.86%, 48.33% and 48.28% in the BI-RADS 2, 3 and 5 classes, respectively. This measure of the BDL model is 73.33%, 59.60%, 67.33% and 59.26% in the BI-RADS 2, 3, 4 and 5 classes, respectively.

- It seems that the misclassification of the BI-RADS category occurs in the radiologist or the model prediction, when there are overlapping the morphological features of benign and malignant lesions extracted using mammography.
- On the other hand, while radiologists would not be able to assign any other BI-RADS scores in some cases and they give them BI-RADS 0, the model was able to distinguish malignant from benign samples with an accuracy of 75.86%. Also, the model was able to correctly identify all malignant samples as BI-RADS 5 in this category.
- Also, the distribution of assigned BI-RADS scores by the radiologist and the model in the benign and malignant samples is examined. The results show that the model tends to use fewer BI-RADS 4 score than the radiologist.
- The similarity comparison of the predicted BI-RADS scores by the radiologist versus the model is also investigated. The investigation shows that although the BI-RADS predictions of the model and the radiologist differ in many samples, there is a great similarity in assigning BI-RADS 5 to malignant samples.
- It is important that the model considers important areas of the lesion, namely, the shape, margins and density to assign the BI-RADS score. Therefore, the extracted features from the final convolutional layer are used to generate a heatmap via the Grad-CAM visualization procedure. The Grad-CAM visualization in correctly and incorrectly predicted BI-RADS scores show that the model pays attention to the shape, margins and density. Meanwhile, due to the similarity of the morphological characteristics of benign lesions to malignancy, or vice versa, the model makes mistakes in predicting the type of some lesions and the corresponding BI-RADS score.

In addition to the above results, perhaps the most important conclusion that can be expressed from this study is that the uncertainty-aware Bayesian Deep Learning model can infer like a radiologist and report its uncertainty about the malignancy of a lesion based on morphological features. This could somewhat indicate the relationship between the AI model and human uncertainty, although in some instances they were different and in others they were similar.

There are some limitations in this study. The limitations of our study are presented below:

1. The number of samples was limited. If the number of the training samples were larger, the model could encounter and learn from different types of lesions with different morphological features. Hence, a fairer comparison between the radiologist and the model could be made.
2. The lack of a medical team did not give us the opportunity to examine the clinical efficacy of the model in assigning BI-RADS score to the mammography images.
3. We just examined the uncertainty information extracted from a selected model, namely VGG16, which performed better than the others.
4. Only one of the uncertainty quantification methods, namely Monte Carlo Dropout as a Bayesian approximation, is used while there are many more methods in the literature of uncertainty quantification in deep neural networks.

Future works will attempt to enhance the performance of the uncertainty based BI-RADS predictor using the bayesian deep learning models with more training data. Also, the uncertainty information extracted from different uncertainty quantification methods will be investigated. In addition, the effectiveness of the model should also be evaluated in a clinical setting by physicians and radiologists.